# On multi-view feature learning

**Roland Memisevic**  RO@CS.UNI-FRANKFURT.DE
University of Frankfurt, Robert-Mayer-Str. 10, 60325 Frankfurt, Germany

## Abstract

Sparse coding is a common approach to learning local features for object recognition. Recently, there has been an increasing interest in learning features from spatio-temporal, binocular, or other multi-observation data, where the goal is to encode the relationship between images rather than the content of a single image. We provide an analysis of multi-view feature learning, which shows that hidden variables encode transformations by detecting rotation angles in the eigenspaces shared among multiple image warps. Our analysis helps explain recent experimental results showing that transformation-specific features emerge when training complex cell models on videos. Our analysis also shows that transformation-invariant features can emerge as a by-product of learning representations of transformations.

## 1. Introduction

Feature learning (AKA dictionary learning, or sparse coding) has gained considerable attention in computer vision in recent years, because it can yield image representations that are useful for recognition. However, although recognition is important in a variety of tasks, a lot of problems in vision involve the encoding of the *relationship* between observations not single observations. Examples include tracking, multi-view geometry, action understanding or dealing with invariances.

A variety of multi-view feature learning models have recently been suggested as a way to learn features that encode relations between images. The basic idea behind these models is that hidden variables sum over products of filter responses applied to *two* observations $\boldsymbol{x}$ and $\boldsymbol{y}$ and thereby correlate the responses.

Adapting the filters based on synthetic transformations on images was shown to yield transformation-specific features like phase-shifted Fourier components when training on shifted image pairs, or "circular" Fourier components when training on rotated image pairs (Memisevic & Hinton, 2010). Task-specific filter-pairs emerge when training on natural transformations, like facial expression changes (Susskind et al., 2011) or natural video (Taylor et al., 2010), and they were shown to yield state-of-the-art recognition performance in these domains. Multi-view feature learning models are also closely related to energy models of complex cells (Adelson & Bergen, 1985), which, in turn, have been successfully applied to video understanding, too (Le et al., 2011). They have also been used to learn within-image correlations by letting input and output images be the same (Ranzato & Hinton, 2010; Bergstra et al., 2010).

Common to all these methods is that they deploy *products of filter responses* to learn relations. In this paper, we analyze the role of these multiplicative interactions in learning relations. We also show that the hidden variables in a multi-view feature learning model represent transformations by detecting rotation angles in eigenspaces that are shared among the transformations. We focus on image transformations here, but our analysis is not restricted to images.

Our analysis has a variety of practical applications, that we investigate in detail experimentally: **(1)** We can train complex cell and energy models using conditional sparse coding models and vice versa, **(2)** It is possible to extend multi-view feature learning to model sequences of three or more images instead of just two, **(3)** It is mandatory that hidden variables pool over multiple subspaces to work properly, **(4)** Invariant features can be learned by *separating* pooling within subspaces from pooling across subspaces. Our analysis is related to previous investigations of energy models and of complex cells (for example, (Fleet et al., 1996; Qian, 1994)), and it extends this line of work to more general transformations than local translation.

## 2. Background on multi-view sparse coding

Feature learning[1] amounts to encoding an image patch $x$ using a vector of latent variables $z = \sigma(W^T x)$, where each column of $W$ can be viewed as a linear feature ("filter") that corresponds to one hidden variable $z_k$, and[2] where $\sigma$ is a non-linearity, such as the sigmoid $\sigma(a) = (1 + \exp(a))^{-1}$. To adapt the parameters, $W$, based on a set of example patches $\{x^\alpha\}$ one can use a variety of methods, including maximizing the average sparsity of $z$, minimizing a form of reconstruction error, maximizing the likelihood of the observations via Gibbs sampling, and others (see, for example, (Hyvärinen et al., 2009) and references therein).

To obtain hidden variables $z$, that encode the *relationship* between two images, $x$ and $y$, one needs to represent correlation patterns between two images instead. This is commonly achieved by computing the sum over products of filter responses:

$$z = W^T (U^T x) * (V^T y) \quad (1)$$

where "$*$" is element-wise multiplication, and the columns of $U$ and $V$ contain image filters that are learned along with $W$ from data (Memisevic & Hinton, 2010). Again, one may apply an element-wise non-linearity to $z$. The hidden units are "multi-view" variables that encode transformations not the content of single images, and they are commonly referred to as "mapping units".

Training the model parameters, $(U, V, W)$, can be achieved by minimizing the *conditional* reconstruction error of $y$ keeping $x$ fixed or vice versa (Memisevic, 2011), or by conditional variants of maximum likelihood (Ranzato & Hinton, 2010; Memisevic & Hinton, 2010). Training the model on transforming random-dot patterns yields transformation-specific features, such as phase-shifted Fourier features in the case of translation and circular harmonics in the case of rotation (Memisevic & Hinton, 2010; Memisevic, 2011). Eq. 1 can also be derived by factorizing the parameter tensor of a conditional sparse coding model (Memisevic & Hinton, 2010). An illustration of the model is shown in Figure 1 (a).

---

[1] We use the terms "feature learning", "dictionary learning" and "sparse coding" synonymously in this paper. Each term tends to come with a slightly different meaning in the literature, but for the purpose of this work the differences are negligible.

[2] In practice, it is common to add constant bias terms to the linear mapping. In the following, we shall refrain from doing so to avoid cluttering up the derivations. We shall instead think of data and hidden variables as being in "homogeneous notation" with an extra, constant 1-dimension.

### 2.1. Energy models

Multi-view feature learning is closely related to *energy models* and to models of *complex cells* (Adelson & Bergen, 1985; Fleet et al., 1996; Kohonen; Hyvärinen & Hoyer, 2000). The activity of a hidden unit in an energy model is typically defined as the sum over squared filter responses, which may be written

$$z = W^T (B^T x) * (B^T x) \quad (2)$$

where $B$ contains image filters in its columns. $W$ is usually constrained such that each hidden variable, $z_k$, computes the sum over only a subset of all products. This way, hidden variables can be thought of as encoding the norm of a projection of $x$ onto a subspace[3]. Energy models are also referred to as "subspace" or "square-pooling" models.

For our analysis, it is important to note that, when we apply an energy model to the *concatenation of two images*, $x$ and $y$, we obtain a response that is closely related to the response of a multi-view sparse coding model (cf., Eq. 1): Let $b_f$ denote a single column of matrix $B$. Furthermore, let $u_f$ denote the part of the filter $b_f$ that gets applied to image $x$, and let $v_f$ denote the part that gets applied to image $y$, so that $b_f^T[x; y] = u_f^T x + v_f^T y$. Hidden unit activities, $z_k$, then take the form

$$z_k = \sum_f W_{fk}(u_f^T x + v_f^T y)^2 = 2 \sum_f W_{fk}(u_f^T x)(v_f^T y) + \sum_f W_{fk}(u_f^T x)^2 + \sum_f W_{fk}(v_f^T y)^2 \quad (3)$$

Thus, up to the quadratic terms in Eq. 3, hidden unit activities are the same as in a multi-view feature learning model (Eq. 1). As we shall discuss in Section 3.5, the quadratic terms do not significantly change the behavior of the hidden units as compared to multi-view sparse coding models. An illustration of the energy model is shown in Figure 1 (b).

## 3. Eigenspace analysis

We now show that hidden variables turn into subspace rotation detectors when the models are trained on transformed image pairs. To simplify the analysis, we shall restrict our attention to transformations, $L$, that are orthogonal, that is, $L^T L = L L^T = I$, where $I$ is the identity matrix. In other words, $L^{-1} = L^T$. Linear transformations in "pixel-space" are also known as *warp*. Note that practically all relevant spatial transformations, like translation, rotation or local shifts, can be expressed approximately as an orthogonal warp, because orthogonal transformations subsume, in particular, all *permutations* ("shuffling pixels").

---

[3] It is also common to apply non-linearities, such as a square-root, to the activity $z_k$.

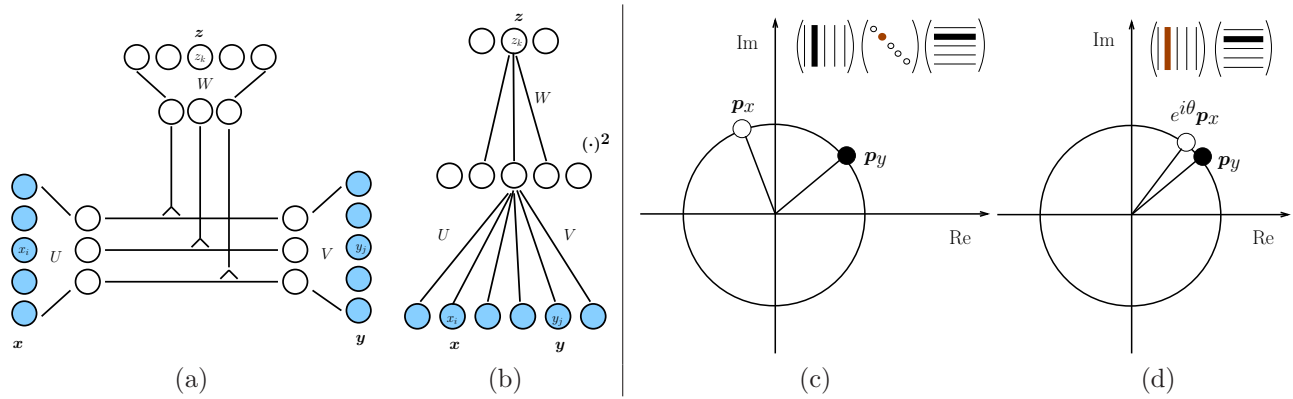

*Figure 1.* **(a)** Modeling an image pair using a gated sparse coding model. **(b)** Modeling an image pair using an energy model applied to the concatenation of the images. **(c)** Projections $p_x$ and $p_y$ of two images onto the complex plane, that is spanned by two eigenfeatures. **(d)** Absorbing eigenvalues into input-features amounts to performing a projection *and* a rotation for image $\boldsymbol{x}$. Hidden units can detect if this brings the projections into alignment (see text for details).

An important fact about orthogonal matrices is that the eigen-decomposition $L = UDU^{\mathrm{T}}$ is complex, where eigenvalues (diagonal of $D$) have absolute value 1 (Horn & Johnson, 1990). Multiplying by a complex number with absolute value 1 amounts to performing a rotation in the complex plane, as illustrated in Figure 1 (c) and (d). Each eigenspace associated with $L$ is also referred to as *invariant subspace* of $L$ (as application of $L$ will keep eigenvectors within the subspace).

Applying an orthogonal warp is thus equivalent to (i) projecting the image onto *filter pairs* (the real and imaginary parts of each eigenvector), (ii) performing a rotation within each invariant subspace, and (iii) projecting back into the image-space. In other words, we can decompose an orthogonal transformation into a set of independent, 2-dimensional rotations. The most well-known examples are translations: A 1D-translation matrix contains ones along one of its secondary diagonals, and it is zero elsewhere[4]. The eigenvectors of this matrix are Fourier-components (Gray, 2005), and the rotation in each invariant subspace amounts to a phase-shift of the corresponding Fourier-feature. This leaves the norm of the projections onto the Fourier-components (the power spectrum of the signal) constant, which is a well known property of translations.

It is interesting to note that the imaginary and real parts of the eigenvectors of a translation matrix correspond to sine and cosine features, respectively, reflecting the fact that Fourier components naturally come in *pairs*. These are commonly referred to as *quadrature pairs* in the literature. The same is true of Gabor features, which represent local translations (Qian, 1994; Fleet et al., 1996). However, the property that eigenvectors come in pairs is not specific to translations. It is shared by all transformations that can be represented by an orthogonal matrix. (Bethge et al., 2007) use the term *generalized quadrature pair* to refer to the eigen-features of these transformations.

### 3.1. Commuting warps share eigenspaces

A central observation to our analysis is that eigenspaces can be *shared* among transformations. When eigenspaces are shared, then the only way in which two transformations differ, is in the angles of rotation within the eigenspaces. In this case, we can represent *multiple transformations with a single set of features* as we shall show.

An example of a shared eigenspace is the Fourier-basis, which is shared among translations (Gray, 2005). Less obvious examples are Gabor features which may be thought of as the eigenbases of local translations, or features that represent spatial rotations. Formally, a set of matrices share eigenvectors if they *commute* (Horn & Johnson, 1990). This can be seen by considering any two matrices $A$ and $B$ with $AB = BA$ and with $\lambda, v$ an eigenvalue/eigenvector pair of $B$ with multiplicity one. It holds that $BAv = ABv = \lambda Av$.

---

[4]To be exactly orthogonal it has to contain an additional one in another place, so that it performs a rotation with wrap-around.

Therefore, $Av$ is also an eigenvector of $B$ with the same eigenvalue.

### 3.2. Extracting transformations

Consider the following task: Given two images $x$ and $y$, determine the transformation $L$ that relates them, assuming that $L$ belongs to a given class of transformations.

The importance of commuting transformations for our analysis is that, since they share an eigenbasis, any two transformations differ only in the angles of rotation in the joint eigenspaces. As a result, one may extract the transformation from the given image pair $(x, y)$ simply by recovering the angles of rotation between the projections of $x$ and $y$ onto the eigenspaces. To this end, consider the real and complex parts $v_R$ and $v_I$ of some eigen-feature $v = v_R + iv_I$, where $i = \sqrt{-1}$. The real and imaginary coordinates of the projection $p_x^v$ of $x$ onto the invariant subspace associated with $v$ are given by $v_R^T x$ and $v_I^T x$, respectively. For the projection $p_y^v$ of the output image onto the invariant subspace, they are $v_R^T y$ and $v_I^T y$.

Let $\phi_x$ and $\phi_y$ denote the angles of the projections of $x$ and $y$ with the real axis in the complex plane. If we normalize the projections to have unit norm, then the cosine of the angle between the projections, $\phi_y - \phi_x$, may be written

$$\cos(\phi_y - \phi_x) = \cos\phi_y \cos\phi_x + \sin\phi_y \sin\phi_x$$

by a trigonometric identity. This is equivalent to computing the inner product between two normalized projections (cf. Figure 1 (c) and (d)). In other words, to estimate the (cosine of) the angle of rotation between the projections of $x$ and $y$, we need to *sum over the product of two filter responses*.

### 3.3. The subspace aperture problem

Note, however, that normalizing each projection to 1 amounts to dividing by the sum of squared filter responses, an operation that is highly unstable if a projection is close to zero. This will be the case, whenever one of the images is almost orthogonal to the invariant subspace. This, in turn, means that the rotation angle *cannot be recovered from the given image*, because the image is too close to the axis of rotation. One may view this as a subspace-generalization of the well-known *aperture problem* beyond translation, to the set of orthogonal transformations. Normalization would ignore this problem and provide the illusion of a recovered angle even when the aperture problem makes the detection of the transformation component impossible. In the next section we discuss how one may overcome this problem by rephrasing the problem as a *detection* task.

### 3.4. Mapping units as rotation detectors

For each eigenvector, $v$, and rotation angle, $\theta$, define the complex output image filter

$$v^\theta = \exp(i\theta)v$$

which represents a projection and simultaneous rotation by $\theta$. This allows us to define a **subspace rotation-detector** with preferred angle $\theta$ as follows:

$$r^\theta = (v_R^T y)(v_R^{\theta\,T} x) + (v_I^T y)(v_I^{\theta\,T} x) \qquad (4)$$

where subscripts $R$ and $I$ denote the real and imaginary part of the filters like before. If projections are normalized to length 1, we have

$$\begin{aligned} r^\theta &= \cos\phi_y \cos(\phi_x + \theta) + \sin\phi_y \sin(\phi_x + \theta) \\ &= \cos(\phi_y - \phi_x - \theta), \end{aligned} \qquad (5)$$

which is maximal whenever $\phi_y - \phi_x = \theta$, thus when the observed angle of rotation, $\phi_y - \phi_x$, is equal to the preferred angle of rotation, $\theta$. However, like before, normalizing projections is not a good idea because of the subspace aperture problem. We now show that mapping units are well-suited to detecting subspace rotations, if a number of conditions are met.

If features and data are *contrast normalized*, then the projections will depend only on how well the image pair represents a given subspace rotation. The value $r^\theta$, in turn, will depend (a) on the transformation (via the subspace angle) and (b) on the content of the images (via the angle between each image and the invariant subspace). Thus, the output of the detector factors in both, the presence of a transformation and our ability to discern it.

The fact that $r^\theta$ depends on image content makes it a suboptimal representation of the transformation. However, note that $r^\theta$ is a "conservative" detector, that takes on a large value only if an input image pair $(x, y)$ complies with its transformation. We can therefore define a content-independent representation by *pooling* over multiple detectors $r^\theta$ that represent the same transformation but respond to different images.

Therefore, by stacking eigenfeatures $v$ and $v^\theta$ in matrices $U$ and $V$, respectively, we may define the representation $t$ of a transformation, given two images $x$ and $y$, as

$$t = W^T P \left(U^T x\right) * \left(V^T y\right) \qquad (6)$$

where $P$ is a band-diagonal within-subspace pooling matrix, and $W$ is an appropriate across-subspace pooling matrix that supports content-independence.

Furthermore, the following conditions need to be met: (1) Images $x$ and $y$ are contrast-normalized, (2) For each row $u_f$ of $U$ there exists $\theta$ such that the corresponding row $v_f$ of $V$ can be written $v_f = \exp(i\theta)u_f$. In other words, filter pairs are related through rotations only.

Eq. 6 takes the same form as inference in a multi-view feature learning model (cf., Eq. 1), if we absorb the within-subspace pooling matrix $P$ into $W$. *Learning* amounts to identifying both the subspaces and the pooling matrix, so training a multi-view feature learning model can be thought of as performing multiple simultaneous diagonalizations of a set of transformations. When a dataset contains more than one transformation class, learning involves partitioning the set of orthogonal warps into commutative subsets and simultaneously diagonalizing each subset. Note that, in practice, complex filters can be represented by learning two-dimensional subspaces in the form of filter pairs. It is uncommon, albeit possible, to learn actually complex-valued features in practice.

It is interesting to note that condition (2) above implies that filters are normalized to have the same lengths. Imposing a norm constraint has been a common approach to stabilizing learning (Ranzato & Hinton, 2010; Memisevic, 2011; Susskind et al., 2011), but it has not been clear why imposing norm constraints help. Pooling over multiple subspaces may, in addition to providing content-independent representations, also help deal with edge effects and noise, as well as with the fact that learned transformations may not be exactly orthogonal. In practice, it is also common to apply a sigmoid non-linearity after computing mapping unit activities, so that the output of a hidden variable can be interpreted as a probability.

Note that diagonalizing a single transformation, $L$, would amount to performing a kind of canonical correlations analysis (CCA), so learning a multi-view feature learning model may be thought of as performing multiple canonical correlation analyzes with tied features. Similarly, modeling within-image structure by setting $x = y$ (Ranzato & Hinton, 2010) would amount to learning a PCA mixture with tied weights. In the same way that neural networks can be used to implement CCA and PCA up to a linear transformation, the result of training a multi-view feature learning model is a simultaneous diagonalization only up to a linear transformation.

### 3.5. Relation to energy models

By concatenating images $x$ and $y$, as well as filters $v$ and $v^\theta$, we may approximate the subspace rotation detector (Eq. 4) with the response of an energy detector:

$$\begin{aligned}
r^\theta &= \left((v_R^\mathrm{T} y) + (v_R^{\theta\,\mathrm{T}} x)\right)^2 + \left((v_I^\mathrm{T} y) + (v_I^{\theta\,\mathrm{T}} x)\right)^2 \\
&= 2\left((v_R^\mathrm{T} y)(v_R^{\theta\,\mathrm{T}} x) + (v_I^\mathrm{T} y)(v_I^{\theta\,\mathrm{T}} x)\right) \\
&\quad + (v_R^\mathrm{T} y)^2 + (v_R^{\theta\,\mathrm{T}} x)^2 + (v_I^\mathrm{T} y)^2 + (v_I^{\theta\,\mathrm{T}} x)^2
\end{aligned} \quad (7)$$

Eq. 7 is equivalent to Eq. 4 up to the four quadratic terms. The four quadratic terms are equal to the sum of the squared norms of the projections of $x$ and $y$ onto the invariant subspace. Thus, like the norm of the projections, they contribute information about the discernibility of transformations. This makes the energy response depend more on the alignment of the images with its subspace. However, like for the inner product detector (Eq. 4), the peak response is attained when both images reside within the detector's subspace and when their projections are rotated by the detectors preferred angle $\theta$.

By pooling over multiple rotation detectors, $r^\theta$, we obtain the equivalent of an energy response (Eq. 3). This shows that energy models applied to the concatenation of two images are well-suited to modeling transformations, too. It is interesting to note that both, multi-view sparse coding models (Taylor et al., 2010) and energy models (Le et al., 2011) were recently shown to yield highly competitive performance in action recognition tasks, which require the encoding of motion in videos.

## 4. Experiments

### 4.1. Learning quadrature pairs

Figure 2 shows random subsets of input/output filter pairs learned from rotations of random dot images (top plot), and from a mixed dataset, consisting of random rotations and random translations (bottom plot). We separate the two layers of pooling, $W$ and $P$, and we constrain $P$ to be band-diagonal with entries $P_{i,i} = P_{i,i+1} = 1$ and 0 elsewhere. Thus, filters need to come in pairs, which we expect to be approximately in quadrature (each pairs spans the subspace associated with a rotation detector $r^\theta$). Figure 2 shows that this is indeed the case after training. Here, we use a modification of a higher-order autoencoder (Memisevic, 2011) for training, but we expect simi-lar results to hold for other multi-view models[5]. We discuss an application of separating pooling in detail in Section 5. Note that for the mixed dataset, both the set of rotations and the set of translations are sets of commuting warps (up to edge effects), but rotations do *not* commute with translations and vice versa. The figure shows that the model has learned to separate out the two types of transformation by devoting a subset of filters to encoding rotations and another subset to modeling translations.

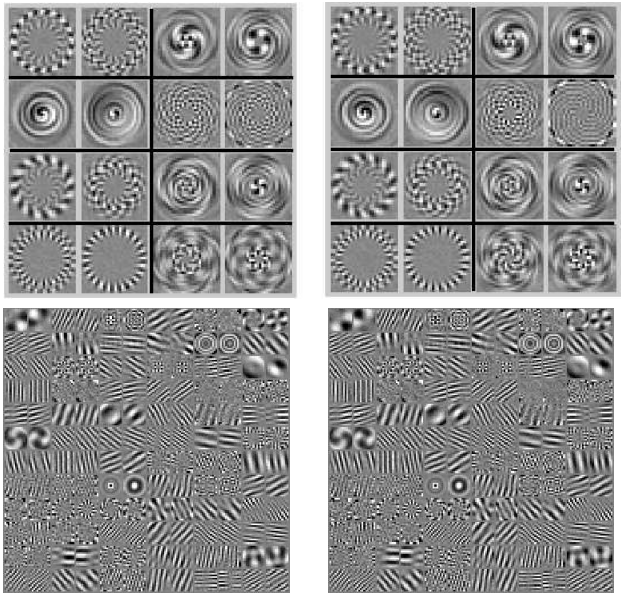

*Figure 2.* Learned quadrature filters. **Top:** Filters learned from rotated images. **Bottom:** Filters learned from images showing both rotations and translations.

### 4.2. Learning "eigenmovies"

Both energy models and cross-correlation models can be applied to more than two images: Eq. 4 may be modified to contain all cross-terms, or all the ones that are deemed relevant (for example, adjacent frames, which would amount to a "Markov"-type gating model of a video). Alternatively, for the energy mechanism, we can compute the square of the concatenation of more than two images in place of Eq. 7, in which case, we obtain the detector response

$$r = \Big(\sum_s {v_R^s}^T x_s\Big)^2 + \Big(\sum_s {v_I^s}^T x_s\Big)^2$$
$$= \Omega + \sum_{st} \big({v_R^s}^T x_s\big)\big({v_R^t}^T x_t\big) + \sum_{st} \big({v_I^s}^T x_s\big)\big({v_I^t}^T x_t\big) \quad (8)$$

---
[5]Code and datasets are available at http://www.cs.toronto.edu/~rfm/code/multiview/index.html

where $\Omega$ contains the quadratic terms of the energy model. In analogy to Section 3.4, for the detector to function properly, features will need to satisfy $v_R^s = \exp(i\theta s) v_R$ and $v_I^s = \exp(i\theta s) v_I$ for appropriate filters $v_R$ and $v_I$.

We verify that training on videos leads to filters which approximately satisfy this condition as follows: We use a gated autoencoder where we set $U = V$, and we set $x = y$ to the concatenation of the 10 frames. In contrast to Section 4.1, we use a single (full) pooling matrix $W$. Figure 3 shows subsets of learned filters after training the model on shifted random dots (top) and natural movies cropped from the van Hateren database (van Hateren & Ruderman, 1998) (center). The learned filter-sequences represent repeated phase-shifts as expected. Thus, they form the "eigenmovies" of each respective transformation class.

Eq. 8 implies that learning videos requires consistency between the filters across time. Each factor – corresponding to a sequence of T filters – can model only the repeated application of *the same* transformation. An inhomogeneous sequence that involves multiple different types of transformation can only be modeled by devoting separate filter sets to homogeneous subsequences. We verify that this is what happens during training, by using 10-frame videos showing random dots that first rotate at a constant speed for 5 frames, then translate at a constant speed for the remaining 5 frames. Orientation, speed and direction vary across movies. We trained a gated autoencoder like in the previous experiment. The bottom plot in Figure 3 shows that the model learns to *decompose* the movies into (a) Fourier filters which are quiet in the first half of a movie and (b) rotation features which are quiet in the second half of the movie.

## 5. Learning invariances by learning transformations

Our analysis suggests that detector responses, $r^\theta$, will not be affected by transformations they are tuned to, as these will only cause a rotation within the detector's subspace, leaving the norm of the projections unchanged. Any *other* transformation (like showing a completely different image pair), however, may change the representation: Projections may get larger or smaller as the transformation changes the degree of alignment of the images with the invariant subspaces.

This suggests, that we can learn features that are *invariant* with respect to one type of transformation and at the same time *selective* with respect to any other type of transformation as follows: we separate the two pooling-levels into a band-matrix $P$ and a full matrix $W$ (cf., Section 4.1). After training the model on a transformation class, the first-level pooling activities $P^{\mathrm{T}}\left(U^{\mathrm{T}}\boldsymbol{x}\right)*\left(V^{\mathrm{T}}\boldsymbol{y}\right)$ computed from a test image pair $(\boldsymbol{x},\boldsymbol{y})$ will constitute a transformation invariant code for this pair. Alternatively, we can use $P^{\mathrm{T}}\left(U^{\mathrm{T}}\boldsymbol{x}\right)*\left(V^{\mathrm{T}}\boldsymbol{x}\right)$ if the test data does not come in the form of pairs but consists only of single images. In this case we obtain a representation of the null-transformation, but it will still be invariant.

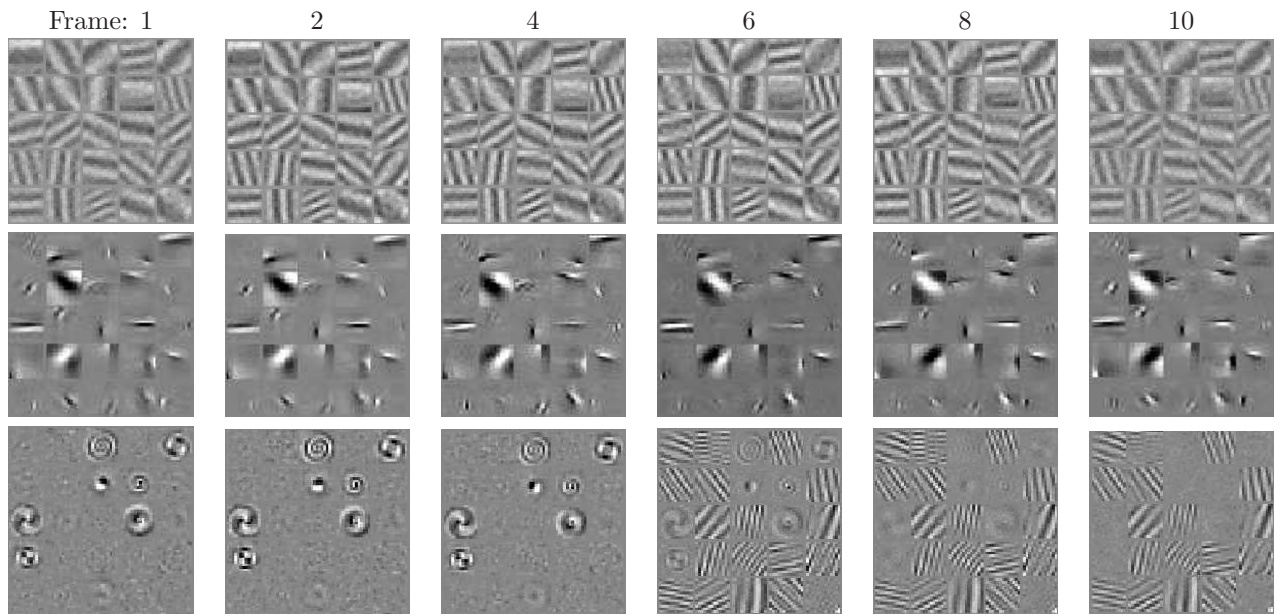

*Figure 3.* Subsets of filters showing six frames from the "eigenmovies" of synthetic movies showing translating random-dots (top); natural movies, cropped from the van Hateren broadcast TV database (middle); synthetic movies showing random-dots which rotate for 5 frames then translate for 5 frames (bottom).

We tested this approach on the "rotated MNIST"-dataset from (Larochelle et al., 2007), which consists of 72000 MNIST digit images of size $28 \times 28$ pixels that are rotated by arbitrary random angles ($-180$ to 180 degrees; 12000 train-, 60000 test-cases, classes range from $0 - 9$). Since the number of training cases is fairly large, most exemplars are represented at most angles, so even linear classifiers perform well (Larochelle et al., 2007). However, when reducing the number of training cases, the number of potential matches for any test case dramatically reduces, so classification error rates become much worse when using raw images or standard features.

We used a gated auto-encoder with 2000 factors and 200 mapping units, which we trained on image pairs showing rotating random-dots. Figure 4 shows the error rates when using subspace features (responses of the first layer pooling units) with subspace dimension 2. We used the features in a logistic regression classifier vs. k-nearest neighbors on the original images (we also tried logistic regression on raw images and logistic regression as well as nearest neighbors on 200-dimensional PCA-features, but the performance is worse in all these cases). The learned features are similar to the features shown in Figure 2, so they are not tuned to digits, and they are not trained discriminatively. They nevertheless consistently perform about as well as, or better than, nearest neighbor. Also, even at half the original dataset size, the subspace features still attain about the same classification performance as raw images on the whole training set. All parameters were set using a fixed hold-out set of 2000 images. The experiment shows that the rotation detectors, $r^\theta$, are affected sufficiently by the aperture problem, such that they are *selective* to image content while being *invariant* to rotation. This shows that we can "harness the aperture problem" to learn invariant features.

## 6. Conclusions

We analyzed multi-view sparse coding models in terms of the joint eigenspaces of a set of transformations. Our analysis helps understand why Fourier features and circular Fourier features emerge when training transformation models on shifts and rotations, and why square-pooling models work well in action and motion recognition tasks. Our analysis furthermore shows how the aperture problem implies that we can learn invariant features as a by-product of learning about transformations.

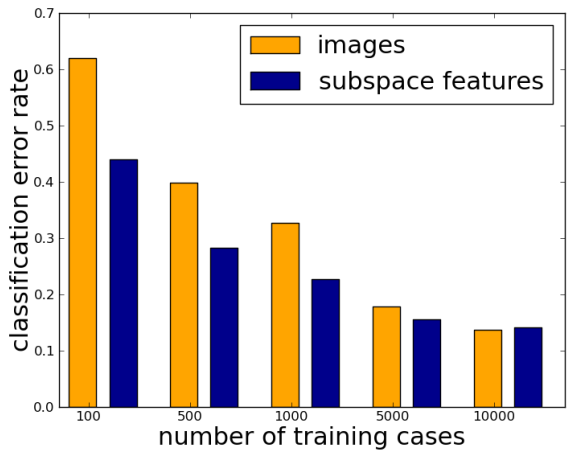

Figure 4. Classification error rate as a function of training set size on the "rotated mnist" dataset, using raw images vs. subspace features trained on rotations.

The fact that squaring nonlinearities and multiplicative interactions can support the learning of relations suggests that these may help increase the role of statistical learning in vision in general. By learning about relations we may extend the applicability of sparse coding models beyond recognizing objects in static, single images, towards tasks that involve the fusion of multiple views, including inference about geometry.

## Acknowledgments

This work was supported by the German Federal Ministry of Education and Research (BMBF) in the project 01GQ0841 (BFNT Frankfurt).